\pdfoutput=1
\documentclass[11pt]{article}

\usepackage[preprint]{acl}

\usepackage{times}
\usepackage{latexsym}
\usepackage{kotex}
\usepackage{xcolor}
\usepackage{multirow}
\usepackage[T1]{fontenc}
\usepackage[utf8]{inputenc}
\usepackage{microtype}
\usepackage{inconsolata}
\usepackage{graphicx}
\usepackage{booktabs}
\usepackage{amsmath} 
\usepackage{algorithm}
\usepackage{algpseudocode}

\usepackage{amssymb}
\usepackage{pifont}% http://ctan.org/pkg/pifont
\usepackage{enumitem}
\usepackage{array}
\usepackage{url}
\usepackage{longtable}

\title{Def-DTS: Deductive Reasoning for Open-domain Dialogue Topic Segmentation}

\author{Seungmin Lee${}^{1}$, Yongsang Yoo${}^{1,2}$, Minhwa Jung${}^{1,3}$, {\bf Min Song${}^{1,4}$\thanks{Corresponding author}} \\ \\
        \normalsize{\textsuperscript{1}Yonsei University},
        \normalsize{\textsuperscript{2}LOTTE INNOVATE},
        \normalsize{\textsuperscript{3}LG Eletronics},
        \normalsize{\textsuperscript{4}Onoma AI}\\
        \small{\textsuperscript{1}\texttt{\{elplaguister, 4n3mone, minalang, min.song\}@yonsei.ac.kr}} \\
        \small{\textsuperscript{2}\texttt{yongsang.yoo@lotte.net},
        \textsuperscript{3}\texttt{minalang.jung@lge.com},
        \textsuperscript{4}\texttt{min.song@onomaai.com}} \\   
}

\begin{document}
\maketitle

\footnotetext{This is a preprint of the camera-ready version accepted to Findings of the ACL 2025. The official version will appear in the ACL Anthology.}

\begin{abstract}
Dialogue Topic Segmentation (DTS) aims to divide dialogues into coherent segments. 
DTS plays a crucial role in various NLP downstream tasks, but suffers from chronic problems: data shortage, labeling ambiguity, and incremental complexity of recently proposed solutions. 
On the other hand, Despite advances in Large Language Models (LLMs) and reasoning strategies, these have rarely been applied to DTS. 
This paper introduces Def-DTS: Deductive Reasoning for Open-domain Dialogue Topic Segmentation, which utilizes LLM-based multi-step deductive reasoning to enhance DTS performance and enable case study using intermediate result. 
Our method employs a structured prompting approach for bidirectional context summarization, utterance intent classification, and deductive topic shift detection. 
In the intent classification process, we propose the generalizable intent list for domain-agnostic dialogue intent classification. 
Experiments in various dialogue settings demonstrate that Def-DTS consistently outperforms traditional and state-of-the-art approaches, with each subtask contributing to improved performance, particularly in reducing type 2 error. 
We also explore the potential for autolabeling, emphasizing the importance of LLM reasoning techniques in DTS. \footnote{Our code and prompts are publicly available at \url{https://github.com/ElPlaguister/Def-DTS}}
\end{abstract}

\section{Introduction}
\label{sec:introduction}

\begin{figure}[t]
  \includegraphics[width=\columnwidth]{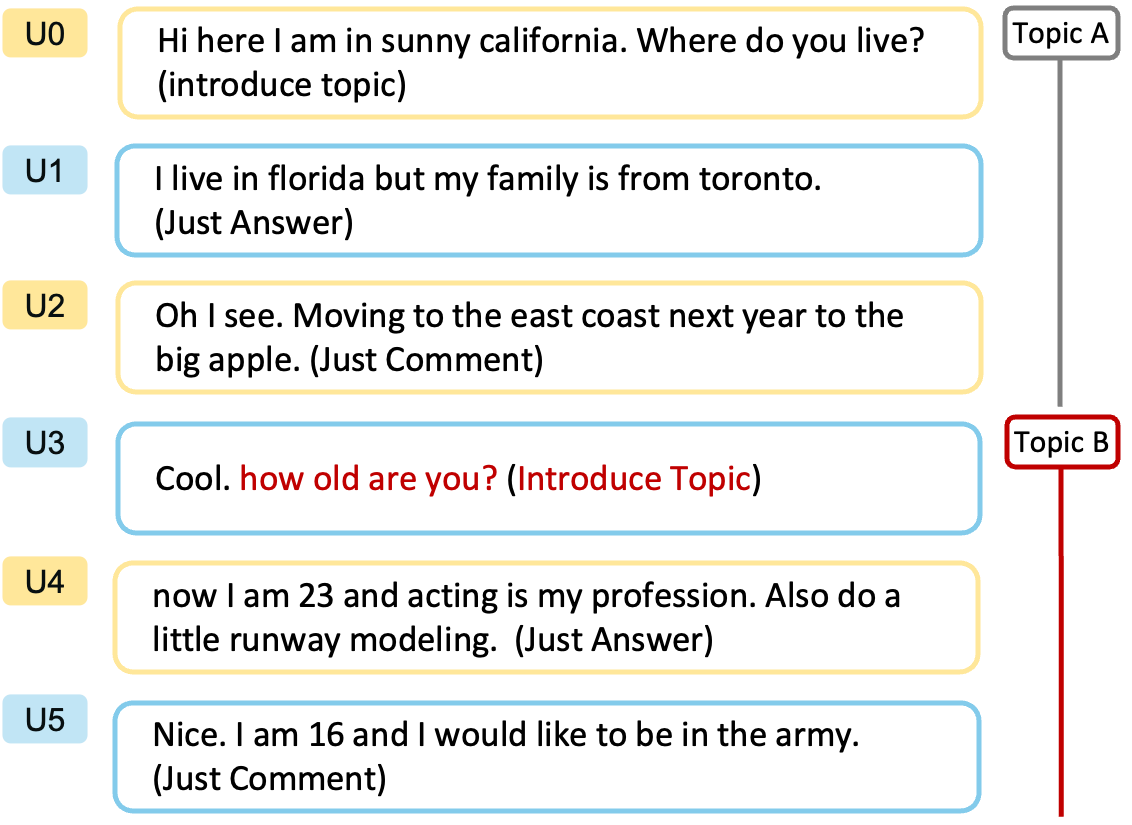}
  \caption{
      An example of a topic shift in a conversation. The cues for a topic shift are highlighted in red.
  }
  \label{fig:dts_example}
\end{figure}

Dialogue Topic Segmentation (DTS) is a task that aims to divide a dialogue into segments where each segment focuses on a coherent topic. 
Figure~\ref{fig:dts_example} shows an example of topic shift within a single dialogue. 
DTS is crucial for various natural language processing (NLP) tasks, including response prediction \citep{resppred1_2020, xu2021topic, resppred2_2022}, response generation \citep{respgen1_2016, respgen2_2021, respgen3_2022}, dialogue state tracking \citep{s3_dst_2024}, summarization \citep{sum1_2016, sum2_2020, sum3_2021, sum4_2022}, question answering \citep{qa1_2018, qa2_2022}, and machine reading comprehension \citep{mrc_seg_2024}.

Despite growing interest, DTS suffers from several chronic challenges. 
First, the shortage of annotated data has led most recent DTS studies to an unsupervised way, which generally yields suboptimal performance. 
Second, ambiguity in segment labeling has hindered the development of effective approaches. 
% Specifically, the low kappa scores of early datasets, such as TIAGE\citep{tiage_2021} reveal inconsistencies in the decisions regarding segment boundaries. 
Lastly, recent studies: DialSTART \citep{dialstart_2023}, UR-DTS \citep{ur_dts_2024} have proposed incremental approaches that require more parameter and complexity, enhancing previous studies: CSM \citep{csm_2021} and DialSTART \citep{dialstart_2023}, respectively. 
This progression indicates that DTS is a challenging and often underestimated problem.

While DTS struggles with its complexities, NLP has witnessed significant advancements with the rise of Large Language Models (LLMs) and reasoning methodologies. 
% LLMs demonstrate remarkable capabilities such as in-context learning, and reasoning strategies like Chain-of-Thought (CoT) have enabled LLMs to handle complex logical tasks and perform factual reasoning using natural language queries. 
However, even considering the robust problem-solving skills of these LLMs and the challenges posed by DTS, reasoning strategies are rarely applied in the DTS area. 
This is because DTS has been treated largely as a lightweight subtask in NLP. 
Nonetheless, with the rise of AI-driven chat services, the demand for more advanced DTS modules is growing. 
LLMs with reasoning capabilities are well suited to meet this need, making LLM-based DTS a viable solution.

To establish clear criteria for topic shifts and simplify complex subtasks, we propose Def-DTS: a deductive reasoning approach for open-domain dialogue topic segmentation using LLM-based multi-step reasoning. 
Def-DTS employs structured prompting to guide LLMs through bidirectional context summarization, utterance intent classification, and deductive topic shift detection at the utterance level, with an emphasis on domain-agnostic intent classification.

To evaluate Def-DTS, we test it on three dialogue datasets spanning open-domain and task-oriented settings across three key metrics. 
Our method consistently outperforms traditional and state-of-the-art unsupervised, supervised, and prompt-based techniques by a significant margin. 
Ablation studies confirm that each subtask improves overall performance, with intermediate intent classification particularly improving true-positive detection. 
Finally, we explore LLM-based auto-labeling for DTS. 
Our contributions are fourfold:
\begin{itemize}
    \item We introduce LLM reasoning techniques to DTS for the first time, consolidating insights from previous methodologies into a coherent and deductive prompt design.
    
    \item We reformulate DTS as an utterance-level intent classification task by implementing intent classification as a core component of a multi-step reasoning process, enabling flexible and task-agnostic prompting.
    
    \item Our method empirically demonstrates superior performance across nearly all comparative baselines, underscoring the efficacy of prompt engineering in DTS.
    
    \item Through an in-depth analysis of our approach's reasoning results, we shed light on the challenges LLM reasoning faces in DTS and discuss the possibility of using LLM as a DTS auto-labeler.
    
\end{itemize}

\section{Related Works}
\label{sec:related_works}

\subsection{Dialogue Topic Segmentation}
\label{sec:related_works_dts}

Dialogue topic segmentation divides dialogues into coherent topic units. 
Due to limited annotated datasets, researchers have largely relied on unsupervised approaches despite their complexity \citep{csm_2021}. Early methods like TextTiling \citep{texttiling_1997} detected topic shifts via lexical similarity, later improved with embedding-based methods \citep{tet_embed_2016}.

Recent research emphasizes topical coherence and similarity scoring. 
CSM \citep{csm_2021} leverages BERT-based coherence, while Dial-START \citep{dialstart_2023} incorporates SimCSE \citep{simcse_2021} for topic similarity.
SumSeg \citep{sumseg_2024} extracts key information via summaries and applies smoothing to handle topic variations. 
UR-DTS \citep{ur_dts_2024} enhances segmentation by rewriting utterances to recover missing references. 
Despite these advances, data scarcity and performance limitations persist.

To address this, SuperDialSeg \citep{superdialseg_2023} introduces a supervised approach using large-scale DGDS datasets \citep{doc2dial_2020, multidoc2dial_2021}. 
However, the available datasets remain limited for open-domain conversations. LLMs are also influencing DTS. S3-DST \citep{s3_dst_2024} applies structured prompting for dialogue state tracking and segmentation, but lacks general applicability to diverse DTS settings.

\subsection{Reasoning Strategy at LLM Inference}
\label{sec:related_works_llm}

The advancement of LLMs \citep{gpt3_2020} has led to research on the integration of System 2 reasoning \citep{kahneman_2011}, including in-context learning \citep{gpt3_2020} and chain-of-thought prompting \citep{cot_2022}. These techniques enable LLMs to tackle complex tasks, such as symbolic mathematics \citep{qwen25math_2024}, retrieval-augmented generation \citep{rag_2020}, and data generation \citep{nemo_2024}. 
Studies show that intermediate reasoning steps significantly enhance performance in areas like multi-hop reasoning \citep{self-prompted_2023} and math problem-solving \citep{mathprompter_2023}.
Building on this, we integrate LLM-based reasoning into DTS, leveraging its potential to improve segmentation accuracy in this inherently complex task.

\section{Def-DTS}
\label{sec:def_dts}

\begin{figure*}[t]
  \includegraphics[width=1.00\linewidth]{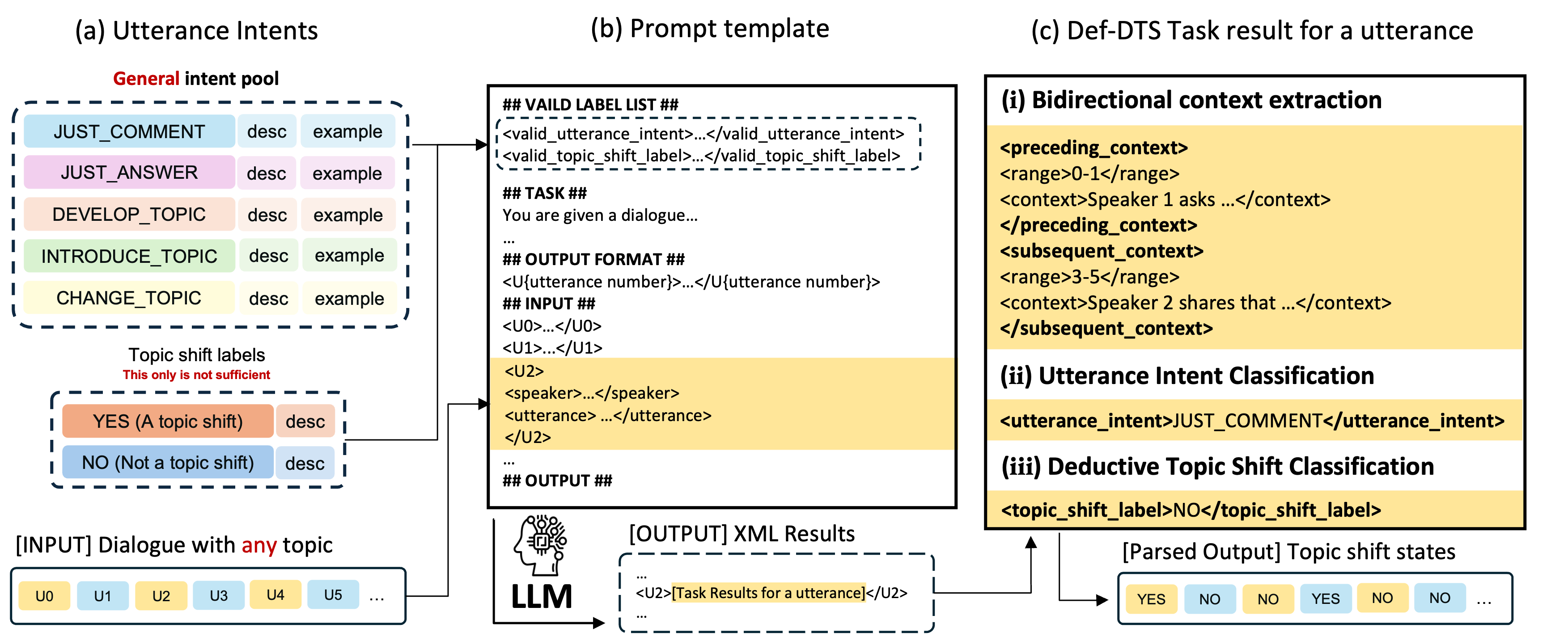}
  \caption {Prompt configuration and overall flow of our method: Def-DTS. (a) We utilize general intent list including the intent-specific examples to enable domain-agnostic categorization. (b) We employ the xml structured input-output format to stably provide the dialogue. (c) We instruct LLM to process the multi-step reasoning for each utterance in a inference.}
  \label{fig:prompt_flow}
\end{figure*}

\subsection{Overall Flow}
\label{sec:def_dts_overall_flow}

Our method (Figure~\ref{fig:prompt_flow}) applies multiple subtasks to each utterance using a structured prompt format. The prompt template (Figure~\ref{fig:prompt_flow}b) includes four parts: valid label list, task description, output format, and input. As shown in Figure~\ref{fig:prompt_flow}c, it comprises three main subtasks: (i) bidirectional context extraction, (ii) utterance intent classification, and (iii) deductive topic shift classification. 
Each subtask is executed deductively, with further details in Algorithm~\ref{fig:ourmethod}.

\begin{algorithm}
\small
\caption{Def-DTS}
\begin{algorithmic}[1]
\Require \textsc{ClassifyIntent}, to classify the intent of an utterance given its context and intent pool  
\Require \textsc{Summarize}, to summarize a given dialogue context
\Require IntentPool $X = \{x_1, x_2, \dots, x_m\}$
\Require StructuredDialogue $D = \{d_i\}_{i=1}^{N}$, where $d_i = \{u_i, s_i\}$ (utterance $u_i$ and speaker $s_i$)
\Ensure Results $R = \{r_i\}_{i=1}^{N}$, where $r_i = \{P_i, Q_i, X_i, T_i\}$ 
\Statex \hspace{\algorithmicindent} (context $P_i, Q_i$, intent $X_i$, topic shift $T_i$)
\State $R \gets \emptyset$, $S \gets \textsc{StructureDialogue}(D)$
\For{$i \gets 1$ to $N$}
    \State $P_i \gets \textsc{ExtractContext}(D, \max(1, i-2), i)$ 
    \State $Q_i \gets \textsc{ExtractContext}(D, i+1, \min(i+3, N))$ 
    \State $X_i \gets \textsc{ClassifyIntent}(D[i], P_i, Q_i, X)$ 
    \State $T_i \gets \textsc{ClassifyTopicShift}(X_i)$
    \State $r_i \gets \{P_i, Q_i, X_i, T_i\}$ 
    \State $R \gets R \cup \{r_i\}$
\EndFor
\State \Return $R$

\Statex
\Function{ClassifyTopicShift}{$x$}
    \State \Return $x \in \{\text{"introduce\_topic"}, \text{"change\_topic"}\}$ ? "YES" : "NO"
\EndFunction

\Statex
\Function{ExtractContext}{$D, start, end$}
    \State $context \gets \textsc{Summarize}(D[start:end])$
    \State \Return $"U_{start}\text{-}U_{end}", context$
\EndFunction

\end{algorithmic}
\label{fig:ourmethod}
\end{algorithm}

\subsection{Structured Format}
\label{sec:def_dts_structured_format}

We benchmark previous DST research \citep{s3_dst_2024} by using a structured template in XML format. Prompt formatting improves adherence to labeling instructions by providing structured and well-defined formats, which improve the alignment with the task and facilitate easier parsing, thereby reducing the need for post-processing \citep{s3_dst_2024}. We believe that our deductive, multi-step method can similarly benefit from these advantages, enhancing alignment with labeling instructions and reducing postprocessing efforts.

To this end, we utilize an XML-based structured prompt to standardize LLM output, enhancing parsing efficiency, and minimizing post-processing. 
The input template (Figure~\ref{fig:prompt_flow}b) organizes utterances in <Ux> elements, each containing the speaker's information and the content of the utterance. 
The output (Figure~\ref{fig:prompt_flow}c) follows the same structured format, ensuring consistent labeling. 
A complete template is available in Appendix~\ref{app:prompt_template}.

\subsection{Bidirectional Context Extraction}
\label{sec:def_dts_bce}

In the first stage of Def-DTS, we instruct the LLM to summarize both the preceding and subsequent dialogues for each utterance. 
Considering bidirectional context is commonly used in many methods such as the BERT architecture \citep{bert_2018} and frequently employed in unsupervised settings \citep{dialstart_2023, ur_dts_2024}, this strategy has proven effective for understanding context in dialogue. 
Although \citealp{s3_dst_2024} only extracts the preceding context to prevent contextual forgetting, we improve upon this by extracting both the preceding and subsequent contexts to enable context-aware dialogue topic segmentation and prevent contextual forgetting.

We opt for a fixed window size to ensure applicability in unsupervised and real-world environments without predefined segments, as it offers a robust approach when segment boundaries are not available. 
As shown in Figure~\ref{fig:prompt_flow}c(i), we instruct the model to summarize no more than two preceding turns using the \texttt{<preceding\_context>} tag and no more than three subsequent turns using the \texttt{<subsequent\_context>} tag. 
This window size of -2:-1 and 1:3 was chosen to balance context informativeness and token efficiency, an approach that is similar to the method used in \citep{dialstart_2023}. 
By summarizing each context range, we aim to conserve tokens while maintaining nuanced topic relationships.

Each element is composed of \texttt{<range> … </range>} for defining the summary scope and \texttt{<context>…</context>} for the actual summary. 
Summarizing in this manner ensures that we capture relevant dialogue while keeping the context concise. 
Our experiments reveal that this approach effectively facilitates the observation of nuanced topic relationships.

\subsection{Utterance Intent Classification}
\label{sec:def_dts_uic}

\begin{table}[htb]
  \centering
  \small
    \resizebox{0.95\linewidth}{!}{
      \begin{tabular}{|c|c|}
        \toprule
          \textbf{Intent}  & \textbf{Description} \\
        \midrule
          JUST & Commenting on the preceding context \\
          COMMENT & without any asking. Not a topic shift \\
        \midrule
          JUST & Answering preceding utterance. \\
          ANSWER & Not a topic shift \\
        \midrule
          DEVELOP & Developing the conversation to similar \\
          TOPIC & and inclusive sub-topics. Not a topic shift \\
        \midrule
          INTRODUCE & Introducing a relevant but different topic. \\
          TOPIC & A topic shift \\
        \midrule
          CHANGE & Completely changing the topic. \\
          TOPIC & A topic shift \\
        \bottomrule
      \end{tabular}
    }
    \caption{Utterance intent list for open-domain dialogue. Using this list, we categorize the utterance and deductively classify the topic shift label.}
    \label{tab:utterance_intents}
\end{table}

As there are issues with the ambiguity of the DTS datasets, simply providing descriptions for topic classification is insufficient to convey a precise definition of topic changes. 
Therefore, we suggest that classifying the utterance using a well-defined and distinct list of labels, similar to intent classification, would be beneficial for DTS. 
Although most of intent classification tasks have been conducted within the context of Task Oriented Dialogue (TOD) \citep{intent1_2016, intent2_2023, jointICSF_1_2023}, TOD typically involves datasets that are specific to a particular domain, making generalization to other domains or more open-ended conversations challenging.

As a way to address this issue, we find that \citealp{tiage_2021} identified five patterns of conversational responses that are utilized in their annotation guidelines. 
These patterns reflect the natural characteristics of the utterances, enabling intent classification for various forms of dialogue. 
Detailed intent patterns and descriptions are in Table~\ref{tab:utterance_intents}.

Inspired by this research, we instruct model to detect topic change through utterance intent classification. 
Specifically, after the bidirectional context extraction, as shown in Figure~\ref{fig:prompt_flow}c(ii), model classifies the utterance into an intent of the predefined general intent pool (upper box of Figure~\ref{fig:prompt_flow}a), considering previously generated bidirectional context.

Additionally, we enhance the model's understanding of the intents by providing example dialogue for each intent, as shown in the General intent pool in Figure~\ref{fig:prompt_flow}a. 
The intent-specific examples provide the model with a helpful guideline detecting topic changes, allowing it to derive results in the subsequent subtask, deductive topic shift. 
Also in order to prove the significance of our intent labels through statistical frequency analysis from the traditional text segmentation method, which can be found in Section~\ref{sec:ling_test}. 
We observed significant performance improvements across various dialogue settings using this technique, enabling a detailed analysis of each intent.
The details of intent pool construction are presented in Appendices~\ref{app:tiage_conversion}--\ref{app:principles_intent}.

\subsection{Deductive Topic Shift Classification}
\label{sec:def_dts_tsc}

Finally, the model predicts whether the topic is changing based on the deductive guidelines from the previous intent classification result, as shown in Figure~\ref{fig:prompt_flow}c(iii). 
This task is processed in an enforced manner, and the model deductively outputs the predetermined label based on the intent classification results of the previous step, according to the tail of description (i.e. Not a topic shift or A topic shift) for each intent described in Table~\ref{tab:utterance_intents}. 

We explicitly instruct the model to process this step for two reasons. 
First, it simplifies the parsing process, making the task easy to handle. 
Second, it ensures that the model explicitly outputs the result of primary goal, allowing it to stay focused on the main objective(DTS), while working through the subtasks.

\section{Experiments}
\label{sec:experiments}

\subsection{Datasets}
\label{sec:datasets}

\begin{table}[htb]
  \tabcolsep=3.5pt
  \centering
  %\setlength{\belowcaptionskip}{-.5cm}
  %\small
  \normalsize
  \renewcommand{\arraystretch}{0.95}
    \resizebox{0.95\columnwidth}{!}{
      \begin{tabular}{ccccccc}
        \toprule
          \multirow{2}{*}{\textbf{Dataset}} & \multirow{2}{*}{\textbf{\# Sample}} & \multicolumn{3}{c}{\textbf{Utterance per Dial.}} & \multicolumn{2}{c}{\textbf{Segment per Dial.}} \\
        \cmidrule(lr){3-5}\cmidrule(lr){6-7}
          & & avg & min & max & avg & avg.len \\
        \midrule
          TIAGE & 100 & 15.6 & 14 & 16 & 4.2 & 3.8 \\
          SuperDialseg & 1322 & 12.1 & 7 & 19 & 4.0 & 3.0 \\
          Dialseg711 & 711 & 26.2 & 7 & 47 & 4.9 & 5.4 \\
        \bottomrule
      \end{tabular}
    }
  \caption{Statistics of datasets for dialogue topic segmentation.}
  \label{tab:simple_data_statistics}
\end{table}

We evaluated our method on three datasets: TIAGE, SuperDialseg, Dialseg711 to verify the performance of our method in both open-domain and task-oriented settings.
Dataset statistics are presented in Table~\ref{tab:simple_data_statistics}. 

TIAGE \citep{tiage_2021} is the only publicly available dataset with topically segmented daily conversations, derived from PersonaChat \citep{personachat_2018} and designed to model topic shifts in open-domain dialogue. 
SuperDialseg \citep{superdialseg_2023} is a large-scale dialogue segmentation dataset based on document-grounded corpora, offering a framework for identifying segmentation points in document-based dialogues. 
Dialseg711 \citep{xu2021topic} is a real-world dialogue dataset auto-labeled from MultiWOZ \citep{multiwoz_2018} and Stanford Dialog Dataset \citep{stanford_2017}, created by joining dialogues with distinct topics, resulting in clear topical differences and low coherency at segment boundaries due to its synthetic nature.

\subsection{Evaluation Metrics}
\label{sec:metrics}

As early studies \citep{csm_2021, superdialseg_2023, sumseg_2024} did, we leverage P$_k$ error \citep{pk_1997}, WindowDiff (WD) error \citep{wd_2002}, and the f1 score. 
The P$_k$ error is calculated by counting the existence of a misallocated segment with a sliding window of predictions. 
The WD error is calculated by comparing the number of boundaries within the sliding window of gold labels and predictions. 
Note that the lower P$_k$ or WD means the higher performance.

\subsection{Comparison Methods}
\label{sec:baselines}

We propose a sophisticated DTS method based on prompt engineering way and compare its performance with various unsupervised, supervised, and LLM-based methodologies. 
First, we compare our method with a random baseline that arbitrarily assigns segment boundaries based on a randomly chosen number of segments. 
Next, we compare it to notable unsupervised learning methods based on TextTiling like Coherence Scoring Model (CSM) \citep{csm_2021}, DialSTART \citep{dialstart_2023}, SumSeg \citep{sumseg_2024}.
We also compare supervised learning methodologies. 
We selected the basic BERT model\citep{bert_2018}, the advanced and high-performing RoBERTa model\citep{roberta_2019}, RetroTS-T5\citep{tiage_2021} system for our comparative analysis.
Finally, we compare our approach with recently introduced LLM-based methods.
We applied these methods using gpt-4o.\footnote{\url{https://platform.openai.com/docs/models/gpt-4o}} We selected the PlainText prompts performed in SuperDialseg\citep{superdialseg_2023} and the prompt of S3-DST\citep{s3_dst_2024}. 
The details of the methodology utilized for each of the actual comparisons are discussed in the Appendix~\ref{app:details_imp}.

\subsection{Experimental Results}
\label{sec:results}

\begin{table*}[thb]
  \centering
  \small
  \renewcommand{\arraystretch}{1.05}
    \resizebox{0.95\linewidth}{!}{
      \begin{tabular}{|l|ccc|ccc|ccc|}
        \toprule
          \multirow{2}{*}{\textbf{Method}}  & \multicolumn{3}{c|}{\textbf{TIAGE}} & \multicolumn{3}{c|}{\textbf{SuperDialseg}} & \multicolumn{3}{c|}{\textbf{Dialseg711}} \\
        \cmidrule(lr){2-4}\cmidrule(lr){5-7}\cmidrule(lr){8-10}
          & \textbf{P$_k$$\downarrow$} & \textbf{WD$\downarrow$} & \textbf{F1$\uparrow$} & \textbf{P$_k$$\downarrow$} & \textbf{WD$\downarrow$} & \textbf{F1$\uparrow$} & \textbf{P$_k$$\downarrow$} & \textbf{WD$\downarrow$} & \textbf{F1$\uparrow$} \\
        \bottomrule
        \toprule
          \multicolumn{10}{|c|}{\textbf{Unsupervised Learning Methods}} \\
        \midrule
          Random & 
          0.526 & 0.664 & 0.237 & 
          0.494 & 0.649 & 0.266 & 
          0.533 & 0.714 & 0.204 \\
        \midrule
          TextTiling & 
          0.469 & 0.488 & 0.204 & 
          0.441 & 0.453 & 0.388 & 
          0.470 & 0.493 & 0.245 \\
          TextTiling+Glove & 
          0.486 & 0.511 & 0.236 & 
          0.519 & 0.524 & 0.353 & 
          0.399 & 0.438 & 0.436 \\
          CSM & 
          \textbf{0.400} & \textbf{0.420} & \textbf{0.427} & 
          0.462 & 0.467 & 0.381 & 
          0.278 & 0.302 & 0.610 \\
          DialSTART & 
          0.482 & 0.528 & 0.378 & 
          \textbf{0.373} & \textbf{0.412} & \textbf{0.627} & 
          \textbf{0.179} & \textbf{0.198} & \textbf{0.733} \\
          SumSeg & 
          0.482 & 0.496 & 0.075 & 
          0.479 & 0.485 & 0.119 & 
          0.477 & 0.483 & 0.070 \\
        \bottomrule
        \toprule
          \multicolumn{10}{|c|}{\textbf{Supervised Learning Methods}} \\
        \midrule
          BERT & 
          0.418 & 0.435 & 0.124 & 
          0.214 & 0.225 & 0.725 & 
          - & - & - \\
          RoBERTa & 
          \textbf{0.265} & \textbf{0.287} & 0.572 & 
          \color{red}\textbf{0.185} & \color{red}\textbf{0.192} & \color{red}\textbf{0.784} & 
          - & - & - \\
          RetroTS-T5 & 0.280 & 0.317 & \textbf{0.576} & 
          0.227 & 0.237 & 0.733 & 
          - & - & - \\
        \bottomrule
        \toprule
          \multicolumn{10}{|c|}{\textbf{LLM-based Methods}}\\
        \midrule
          Plain Text & 
          0.445 & 0.485 & 0.185 & 
          0.412 & 0.427 & 0.048 & 
          0.333 & 0.353 & 0.010 \\
          S3-DST\textsubscript{uttr} & 
          0.439 & 0.498 & 0.265 & 
          0.442 & 0.469 & 0.404 &    
          0.087 & 0.109 & 0.790 \\
          Def-DTS (Ours) & 
          \color{red}\textbf{0.232} & \color{red}\textbf{0.256} & \color{red}\textbf{0.699} & 
          \textbf{0.315} & \textbf{0.324} & \textbf{0.686} & 
          \color{red}\textbf{0.015} & \color{red}\textbf{0.018} & \color{red}\textbf{0.979} \\
        \bottomrule
      \end{tabular}
      }
  \caption{Performances on three datasets. Due to absence of train and validation split for Dialseg711 dataset, There are no report in dialseg711's supervised learning part. The best results for each method group are highlighted in \textbf{bold}. The best performances around all method are indicated as \color{red}\textbf{red} \color{black}colored text.}
  \label{tab:result}
\end{table*}

The experimental results are presented in Table~\ref{tab:result}. 
Def-DTS consistently demonstrated superior performance among LLM-based methods and achieved state-of-the-art results on the TIAGE and Dialseg711 datasets, which are closely aligned with our objective of analyzing general open-domain dialogues.
In contrast, other LLM-based methods showed lower performance not only compared to supervised learning and some unsupervised methods, indicating that simply using an LLM is not a guarantee of success.

\paragraph{TIAGE}

Compared to S3-DST\textsubscript{uttr}, the recent LLM-based method, our method achieved reductions of more than 0.2 in both P$_k$ and WD errors, along with an impressive increase of more than 0.4 in the F1 score.
% compared to  S3-DST\textsubscript{uttr}. 
Furthermore, Def-DTS outperformed even the supervised approaches in TIAGE, surpassing them in all metrics by over 10\%, thus highlighting the effectiveness of our approach in achieving high performance across various dialogue environments without additional training, even in domain-agnostic settings.

\paragraph{SuperDialseg}

Our method outperformed all unsupervised methods. 
Although it showed lower performance compared to models trained using supervised learning, it achieved the best results among unsupervised methods that use prompt-based techniques.

\paragraph{Dialseg711} 

Our approach also delivered superior performance. Notably, it surpassed the strongest LLM-based approach, S3-DST\textsubscript{uttr}, underscoring the general applicability and robustness of our method.

Overall, these consistent improvements across all tested datasets confirm the robust effectiveness of our approach for diverse open-domain dialogue scenarios, demonstrating that it not only excels in unsupervised settings, but also surpasses previously leading LLM-based methods. This further establishes potential of our method for delivering high performance even under challenging, domain-agnostic conditions.

\section{Analysis and Discussion}
\label{sec:analysis}

\subsection{Ablation Study}
\label{sec:ablation_study}

% SuperDialseg, Dialseg711에 대해 수행 예정
\begin{table}[htb]
  \centering
  % \small
  \tiny
    \resizebox{0.95\linewidth}{!}{
      \begin{tabular}{cccc}
        \toprule
          \multirow{2}{*}{\textbf{Method}}  & \multicolumn{3}{c}{\textbf{TIAGE}} \\
          \cmidrule(lr){2-4}
          & \textbf{P$_k$$\downarrow$} & \textbf{WD$\downarrow$} & \textbf{F1$\uparrow$} \\
        \midrule
          w/o all & 0.295 & 0.333 & 0.605 \\
          w/o intent & 0.316 & 0.342 & 0.524 \\
          w/o examples & 0.287 & 0.308 & 0.617 \\
          w/o context & 0.263 & 0.296 & 0.682 \\
          w/o bidirectional & 0.269 & 0.301 & 0.659 \\
          Def-DTS & \textbf{0.232} & \textbf{0.256} & \textbf{0.699} \\
        \bottomrule
      \end{tabular}
    }
    \caption{Ablation study.}
    \label{tab:ablation_study}
\end{table}

To assess the contributions of each part of our approach, we performed an ablation study. The results are shown in Table~\ref{tab:ablation_study}. 
In the \textbf{w/o all} components case, the model is instructed to detect topic shifts without context extraction or intent classification. In the \textbf{w/o intent} case, the model detects topic shifts after context extraction for each utterance.
We observe that w/o intent performs worse than w/o all components, indicating that relying solely on dialogue context for topic shift prediction does not yield optimal performance.
In the \textbf{w/o examples} case, this is essentially Def-DTS but without examples for intent.
w/o examples performed better than w/o intent, showing that processing context into intent before using it for topic shift prediction provides a significant advantage.
In the \textbf{w/o context} case, the model is instructed to detect topic shifts after intent classification for each utterance, which is the opposite of the w/o intent case.
This result demonstrates that intent classification, when supported by appropriate examples, has a significant impact on topic shift prediction for individual utterances.
In the \textbf{w/o bidirectional} context case, the subsequent context is not considered at the context extraction step.
Compared to both the Def-DTS and w/o context case, this case showed lower performance, highlighting that considering bidirectional context is crucial for intent classification and topic shift detection.
In summary, each module of \textbf{Def-DTS} contributes to performance improvement, and when all modules are applied, they work synergistically to yield a substantial increase in performance.

\subsection{Comparative Study for Structured Format}
\label{sec:comparative_study}

\begin{table}[htb]
  \centering
  \tiny
    \resizebox{0.95\linewidth}{!}{
      \begin{tabular}{cccc}
        \toprule
          \multirow{2}{*}{\textbf{I/O Format}}  & \multicolumn{3}{c}{\textbf{TIAGE}} \\
          \cmidrule(lr){2-4}
          & \textbf{P$_k$$\downarrow$} & \textbf{WD$\downarrow$} & \textbf{F1$\uparrow$} \\
        \midrule
          NL & 0.274 & 0.302 & 0.640 \\
          JSON & 0.259 & 0.292 & 0.658 \\
          XML & 0.232 & 0.256 & 0.699 \\
        \bottomrule
      \end{tabular}
    }
    \caption{Comparative study for structured format.}
    \label{tab:comparative_study}
\end{table}

To examine the impact of structured I/O formats not covered in the ablation study, we represent identical prompts in three different formats—Natural Language (NL), JSON, and XML—and compare their performance in Table~\ref{tab:comparative_study}. The results show that the structured formats, XML and JSON, not only offer parsing advantages but also outperform NL in terms of task performance. These findings empirically support the hypothesis proposed by \citealp{s3_dst_2024} that XML can provide structural benefits in dialogue processing.

\subsection{Intent Classification Accuracy}
\label{sec:analysis2}

\begin{table}[htb]
  \centering
  \small
  \renewcommand{\arraystretch}{1.05}
    \resizebox{\linewidth}{!}{
      \begin{tabular}{|c|ccccc|c|}
        \toprule
          \textbf{Intent} & TP & FP & TN & FN & Acc \\
        \midrule
          JUST\_COMMENT & 0 & 1 & 498 & 35 & 0.93\\
          JUST\_ANSWER & 0 & 1 & 456 & 23 & 0.95 \\
          DEVELOP\_TOPIC & 0 & 0 & 119 & 47 & 0.71 \\
          INTRODUCE\_TOPIC & 189 & 68 & 0 & 0 & 0.73 \\
          CHANGE\_TOPIC & 21 & 6 & 0 & 0 & 0.78 \\
        \bottomrule
      \end{tabular}
      }
    \caption{Intent-level confusion matrix for TIAGE benchmark.}
    \label{tab:n_confusion_matrix}
\end{table}

As our method deduces topic shifts directly from utterance intent, analyzing the intent classification results is crucial. 
We examine the confusion matrix (Table~\ref{tab:n_confusion_matrix}) to identify the utterance types that the model struggles with. 
Since there is no ground truth about intent and only topic shift labels are provided, correctness is determined by whether the predicted intent aligns with the topic shift label. 
For instance, if an utterance is a topic shift but classified as JUST\_COMMENT for intent and NO for topic shift, it counts as a False Negative (FN).

Most results performed well in topic shift classification, except for two cases: positives in JUST\_COMMENT and JUST\_ANSWER. 
These findings indicate that the model's primary challenge lies in distinguishing subtle topic differences as actual shifts, a more significant factor in performance degradation than other utterance types. 
Analysis of additional datasets (SuperDialseg, Dialseg711) is in Appendix~\ref{app:others}.

\subsection{Intent Level Comparison}
\label{sec:analysis3}

\begin{figure}[htb]
  \includegraphics[width=0.55\columnwidth]{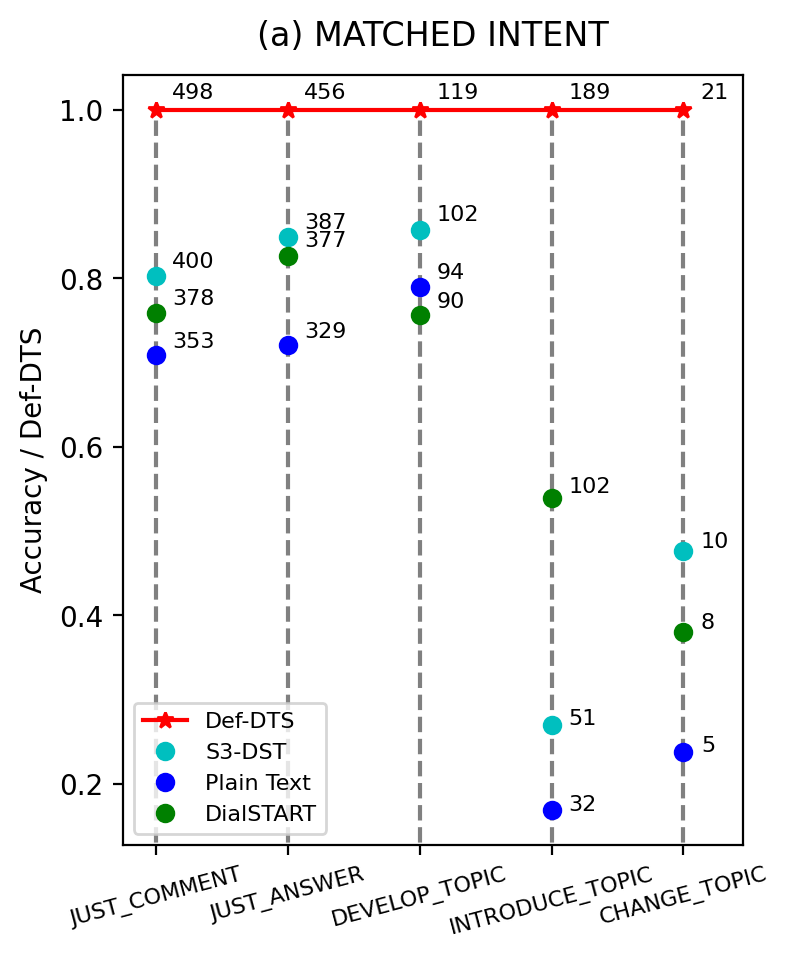} 
  \includegraphics[width=0.44\columnwidth]{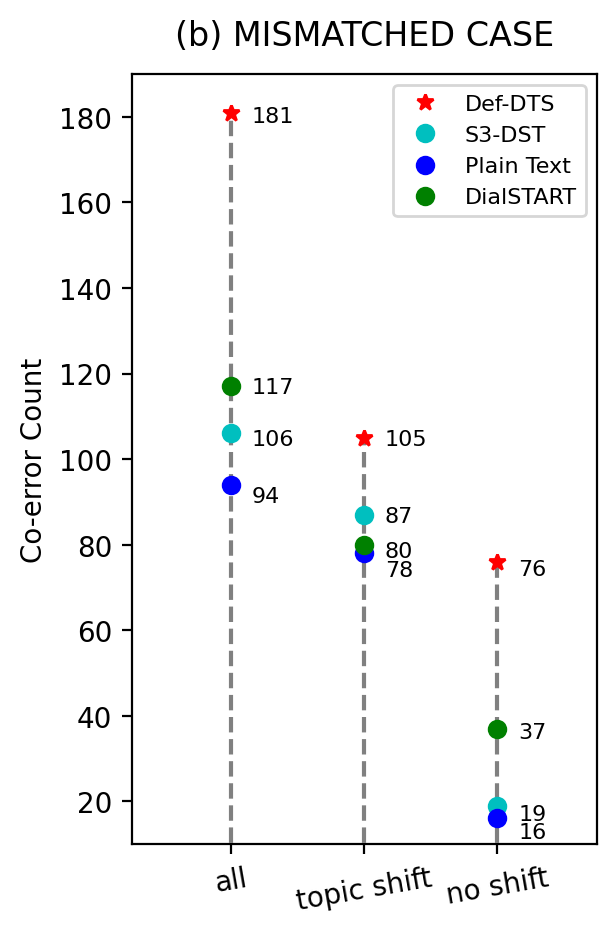}
  \caption {
  (a) MATCHED INTENT indicates the accuracy of the other methodologies for grouped utterances by our intent classification process only in the true cases of our method.
  (b) MISMATCHED CASE indicates the co-error count of the other methodologies with our methods for only in the false cases of our method.
    }
    \label{fig:analysis3}
\end{figure}

We compared the performance of various approaches across different intent categories predicted by our method, as shown in Figure~\ref{fig:analysis3}. 
In (a) MATCHED INTENT, for utterances without a topic shift, other methodologies achieved approximately 80-85\% accuracy when our method was also correct. However, for utterances that involve a topic shift, the accuracy of other methods dropped to around 20-50\% for the correct case of our method. In (b) MISMATCHED CASE, for utterances without a topic shift, other methods correctly classified 50\% of the cases where our method was incorrect. However, for utterances with a topic shift, other methods failed to classify 80\% of the cases that our method also missed. This demonstrates that detecting utterances with actual topic shifts is considerably more challenging than detecting those without topic shifts. Our method outperforms the other methodologies by roughly 20\% in cases without topic shifts, and by over 40\% in cases with topic shifts. In summary, while our approach improves accuracy across all cases, it shows even greater improvement when handling utterances with topic shifts.

\subsection{Linguistic Test for Intent Labels}
\label{sec:ling_test}

To demonstrate the impact of intent labels on topic shifts, we adopted methods from statistical linguistics. Traditional text segmentation uses pauses, cue words, and referential noun phrases to identify boundaries \citep{passonneau-litman-1997-discourse}. 
\citealp{galley-etal-2003-discourse} found a significant correlation between cue phrases and topic segmentation.
Building on this, we hypothesized that cue words in labels like "introduce topic" and "change topic" correlate with their overall frequency in the data. A $\chi^2$ test yielded $\chi^2$(32) = 76.2263, p < 0.001, confirming a significant relationship. This validates our labels as linguistically rich markers of discourse boundaries and provides a criterion for selecting topic-shift data in new datasets. 

\subsection{Performance Comparison for Local LLMs}
\label{sec:analysis4}

\begin{table*}[thb]
  \tabcolsep=3.5pt
  \centering
  \small
  \renewcommand{\arraystretch}{1.05}
    \resizebox{\linewidth}{!}{
      \begin{tabular}{|l|ccc|ccc|ccc|}
        \toprule
          \multirow{2}{*}{\textbf{Model}}  & \multicolumn{3}{c|}{\textbf{TIAGE}} & \multicolumn{3}{c|}{\textbf{SuperDialseg}} & \multicolumn{3}{c|}{\textbf{Dialseg711}} \\
        \cmidrule(lr){2-4}\cmidrule(lr){5-7}\cmidrule(lr){8-10}
          & \textbf{P$_k$$\downarrow$} & \textbf{WD$\downarrow$} & \textbf{F1$\uparrow$} & \textbf{P$_k$$\downarrow$} & \textbf{WD$\downarrow$} & \textbf{F1$\uparrow$} & \textbf{P$_k$$\downarrow$} & \textbf{WD$\downarrow$} & \textbf{F1$\uparrow$} \\
        \bottomrule
        \midrule
          Plain Text + Llama & 
          0.472 & 0.515 & 0.215 &
          0.492 & 0.495 & 0.026 & 
          0.350 & 0.373 & 0.032 \\
          Plain Text + Qwen & 
          0.495 & 0.533 & 0.162 &
          0.485 & 0.487 & 0.059 &
          0.422 & 0.434 & 0.012 \\
        \midrule
          S3-DST\textsubscript{uttr} + Llama & 
          0.456 & 0.474 & 0.143 &
          0.490 & 0.512 & 0.072 & 
          0.158 & 0.190 & 0.553 \\
          S3-DST\textsubscript{uttr} + Qwen & 
          - & - & - &
          - & - & - &
          - & - & - \\
        \midrule
          Def-DTS + Llama & 
          \textbf{0.307} & \textbf{0.339} & \textbf{0.552} &
          \textbf{0.384} & \textbf{0.385} & \textbf{0.432} & 
          \textbf{0.029} & \textbf{0.039} & \textbf{0.941} \\
          Def-DTS + Qwen & 
          0.327 & 0.345 & 0.530 &
          0.433 & 0.434 & 0.171 &
          0.102 & 0.208 & 0.729 \\
        \bottomrule
      \end{tabular}
      }
  \caption{Performances on the local LLMs. We employed Llama-3.1-70B-Instruct and Qwen2.5-72B-Instruct for Llama and Qwen, respectively. The case of best performance across all method are highlighted in \textbf{bold}.}
  \label{tab:main_local_llms}
\end{table*}

We conducted experiments using local LLMs instead of GPT-4, specifically using Lama 3.1 and Qwen 2.5, with the exact model names listed in Table~\ref{tab:main_local_llms}. For the experiments, we tested three prompts: Plain Text, S3-DST\textsubscript{uttr}, and Def-DTS. To ensure efficiency, we randomly sampled 100 examples from each dataset. 
Def-DTS achieved the highest performance in all datasets in this experiment. 
In the case of Qwen, formatting errors were observed in all datasets. Although plain is an unstructured method, it did not have errors, but its performance remained comparatively lower. These results demonstrate that Def-DTS maintains high accuracy in different LLMs.
Experiments with more LLMs are presented in the Appendices~\ref{app:local_llms}--~\ref{app:closed_llms}.

\subsection{Discussion for Possibility of Auto-Labeling}
\label{sec:discussion}

As various auto-labeling methodologies have been proposed to date, we assess prompt engineering could potentially serve as a viable auto-labeling methodology.
We conducted a preliminary experiment to assess the feasibility of using prompt engineering for DTS. We compared the segment labels generated by GPT-4 for Def-DTS with the correct labels using Cohen's Kappa score. The results showed Kappa scores of 0.485 for TIAGE, 0.429 for SuperDialseg, and 0.975 for Dialseg711, indicating moderate agreement for TIAGE and SuperDialseg, and almost perfect agreement for Dialseg711. 
Notably, our labeling result for TIAGE exceeded the 0.479 agreement score observed between actual human annotators. 
While improvements are needed given the moderate agreement, these findings suggest that our approach can still function as a minimal annotator.

\section{Conclusion}
\label{sec:conclusion}

Previous approaches to DTS have been constrained by several challenges, including data shortage, ambiguity of segment labeling, and increasingly complex model architectures. 
Concurrently, The promising approach of reasoning with LLMs has yet to be explored in the context of DTS. To address these issues, we propose Def-DTS, that leverages LLMs in conjunction with sophisticated reasoning strategies. Def-DTS incorporates bidirectional context extraction, a crucial component in previous research, along with the novel task of utterance intent classification. This approach demonstrates significant performance improvements in both the open-domain dialogue setting and the task-oriented dialogue setting. Its efficacy across diverse datasets is enhanced through the provision of dataset-specific examples in the utterance intent classification task, enabling adaptable performance in varied dialogue contexts. Through its primary findings and diverse analysis, we demonstrate the efficacy of LLM-reasoning as a promising approach to DTS. It not only highlights the potential of our method, but it also statistically delineates the challenges to be addressed in future research. 
In subsequent investigations, we intend to explore the feasibility of automated labeling for DTS and examine the potential integration of DTS with other NLP downstream tasks through LLM reasoning.

\section{Limitations}
\label{sec:limitations}

Firstly, though we have demonstrated the significance of the current intent labels by statistical linguistic experiments, we cannot entirely rule out the possibility that more suitable intent labels exist. Additionally, as we provide a first approach for selecting representative examples, a more in-depth exploration of methodologies for selecting optimal examples remains a future step of this research.
In order to improve the quality of intents and examples in a variety of dialogue settings, the fundamental problem must be addressed first, namely the provision of quality datasets for DTS. Dialogue should include thorough labeling criteria and realistic dialogue domains to address our limitations. However, human labeling is not only still expensive but also carries the risk of inconsistent or ambiguous labeling. We believe that automated labeling using LLMs with a sophisticated guideline will play a crucial role in creating a more sustainable and reliable DTS environment.

\section*{Acknowledgments}

This research was supported by the Culture, Sports and Tourism R\&D Program through the Korea Creative Content Agency grant funded by the Ministry of Culture, Sports and Tourism in 2024 (Project Name: Developing a generative AI story platform for Fanfiction, Project Number: RS-2024-00442270). This research was partly supported by an IITP grant funded by the Korean Government (MSIT) (No. RS-2020-II201361 , Artificial Intelligence Graduate School Program (Yonsei University))

\bibliography{custom}

\appendix
\onecolumn
\section{Prompts}
\label{app:prompts}

\subsection{Prompt Template}
\begin{longtable}{p{\linewidth}}
\hline
Prompt Template \\
\hline
\endhead
<valid\_utterance\_intent>
<item> \\
<name>JUST\_COMMENT</name> \\
<desc>Commenting on the preceding context without any asking. Not a topic shift</desc> \\
<example> \\
<speaker1>My dad works for the New York Times.</speaker1> \\
<speaker2>Oh wow! You know, I dabble in photography; maybe you can introduce us sometime.</speaker2> \\
<speaker1>Photography is the greatest art out there. (not a topic shift)</speaker1> \\
</example> \\
</item> \\
<item> \\
<name>JUST\_ANSWER</name> \\
<desc>Answering preceding utterance. Not a topic shift</desc> \\
<example> \\
<speaker1>Do you teach cooking? </speaker1> \\
<speaker2>No, since I’m a native of Mexico, I teach Spanish. (not a topic shift)</speaker2> \\
</example> \\
</item> \\
<item> \\
<name>DEVELOP\_TOPIC</name> \\
<desc>Developing the conversation to similar and inclusive sub-topics. Not a topic shift</desc> \\
<example> \\
<speaker1>Pets are cute!</speaker1> \\
<speaker2>I heard that Huskies are difficult dogs to take care of. (not a topic shift)</speaker2> \\
</example> \\
</item> \\
<item> \\
<name>INTRODUCE\_TOPIC</name> \\
<desc>Introducing a relevant but different topic. A topic shift</desc> \\
<example> \\
<speaker1>You are an artist? What kind of art, I do American Indian stuff.</speaker1> \\
<speaker2> I love to eat too, sometimes too much. (a topic shift)</speaker2> \\
</example> \\
</item> \\
<item> \\
<name>CHANGE\_TOPIC</name> \\
<desc>Completely changing the topic. A topic shift</desc> \\
<example> \\
<speaker1>What do you do for fun?</speaker1> \\
<speaker2>I drive trucks so me and my buds go truckin in the mud.</speaker2> \\
<speaker1>Must be fun! My version of that’s running around a library!</speaker1> \\
<speaker2>That's cool! I love that too. Do you have a favourite animal? Chickens are my favourite. I love them. (topic shift)</speaker2> \\
</example> \\
</item> \\
</valid\_utterance\_intent> \\
<valid\_topic\_shift\_label> \\
<item> \\
<name>YES</name> \\
<desc>The current utterance has **weak OR no topical** relation to the preceding conversation context OR is the first utterance in the conversation, marking the beginning of a new dialogue segment.</desc> \\
</item> \\
<item> \\
<name>NO</name> \\
<desc>The current utterance has **relevant OR equal** topic to the preceding conversation context.</desc> \\
</item> \\
</valid\_topic\_shift\_label> \\
 \\
\#\# TASK \#\# \\
You are given a dialogue starting with U. From utterance number 0, you have to answer the following sub-tasks for each utterance. \\
1. Summarize the preceding and subsequent context in <=3 sentences seperately \\
The range of the context should be previous or next 1-3 utterances except for the case of the first or last utterance. \\
For example, given current utterance number is 2, preceding range is 0-1, subsequent range is 3-5. \\
2. Output the utterance\_intent \\
Use the list <valid\_utterance\_intent> \textrm{...} </valid\_utterance\_intent> to categorize utterance. \\
Consider topical difference between preceding and subsequent context. \\
3. Output the topic\_shift\_label \\
Use the list <valid\_topic\_shift\_label> \textrm{...} </valid\_topic\_shift\_label>. \\
 \\
\#\# OUTPUT FORMAT \#\# \\
<U\{utterance number\}> \\
<preceding\_context> \\
<range>\{range of utterances referred in context\}</range> \\
<context>\{context of the previous 1-3 utterances\}</context> \\
</preceding\_context> \\
<subsequent\_context> \\
<range>\{range of utterances referred in context\}</range> \\
<context>\{context of the next 1-3 utterances\}</context> \\
</subsequent\_context> \\
<utterance\_intent>\{valid utterance intent\}</utterance\_intent> \\
<topic\_shift\_label>\{valid topic shift label\}</topic\_shift\_label> \\
</U\{utterance number\}> \\
 \\
\#\# INPUT \#\# \\
\{XML-structured dialogue\} \\
 \\
\#\# OUTPUT \#\# \\
\hline
\caption{We provide prompt template for main dataset: TIAGE.
Each dataset has different characteristics to other datasets, so we modified intent pool from original template for each dataset.}
\label{tab:15}
\end{longtable}
\label{app:prompt_template}

\twocolumn
\subsection{Intent Labels for other datasets}
\label{app:prompt_other_labels}

\begin{table}[htb]
  \centering
  \small
    \resizebox{1.00\linewidth}{!}{
      \begin{tabular}{|c|c|}
        \toprule
          \textbf{Intent}  & \textbf{Description} \\
        \midrule
          DIFFERENT & Questioning about something that is not similar or \\
          QUESTION & topically different to preciding context. A topic shift \\
        \midrule
          RELEVANT & Questioning about something that is similar or topically  \\
          QUESTION & coherent to preceding context. Not a topic shift \\
        \midrule
          \multirow{2}{*}{ANSWERING} & Answering preceding utterance. \\
           & Not a topic shift \\
        \midrule 
          ADDITIONAL & An additional comment from the same speaker  \\
          COMMENT & in addition to a previous utterance. Not a topic shift \\
        \bottomrule
      \end{tabular}
    }
    \caption{Utterance intent list for SuperDialseg dataset.}
    \label{tab:utterance_intents_superseg}
\end{table}

For the Dialseg711 dataset, we delete an intent named INTRODUCE\_TOPIC from the original intent list.
For the SuperDialseg dataset, the topic shift occurs when the utterance refers to a different document to previous utterance. As shown in Table~\ref{tab:utterance_intents_superseg}, we completely change the intent from the original to fit to document grounded dialogue setting based topic transition.

\subsection{Modification of TIAGE Response Patterns}
\label{app:tiage_conversion}

\begin{table}[htb]
    \centering
    \small
    \resizebox{1.00\linewidth}{!}{
    \begin{tabular}{|c|c|}
        \toprule
        \textbf{Scenario} & \textbf{Topic Shift} \\
        \midrule
        Commenting on the previous context & No \\
        Question answering & No \\
        Developing the conversation to sub-topics & No \\
        Introducing a relevant but different topic & Yes \\
        Completely changing the topic & Yes \\
        \bottomrule
    \end{tabular}
    }
    \caption{Intent labels in different dialog scenarios.}
    \label{tab:tiage_intent_labels}
\end{table}

TIAGE \citep{tiage_2021} is originally designed for real-time topic shift detection without using a subsequent context. 
Consequently, its conversation response pattern list cannot be applied as is for our full-dialogue segmentation task. 
To address this, we propose two methods. 
First, instead of classifying each utterance with only its immediate context, we retrieve both preceding and subsequent context to inform our decisions. 
This bidirectional view captures how an utterance relates to what was said before and what follows, enabling more precise intent classification. 
Second, we adapt TIAGE's original response pattern list to our classification objectives by reorganizing patterns (e.g., Asking) and refining the description of each intent (e.g., Relevant, Inclusive). 
This transformation ensures a more granular detection of whether an utterance continues the topic or shifts in a subtle way. 
By employing these enhancements, we achieve significantly better performance in segmenting topics across entire dialogues, surpassing the results obtained by using TIAGE’s list unaltered.

\subsection{Construction of Intent Examples}
\label{app:details_ex}
For the TIAGE dataset, we used the examples directly from their paper.
For other datasets, we randomly selected parts of the conversation from the Train split that adhere to the following rules:
\begin{itemize}
    \item Select 2–3 consecutive utterances for each example.
    \item Ensure that the final utterance in the example corresponds to the target utterance intent.
    \item Extract all the examples from a single dialogue.
    \item Keep the utterance lengths concise (within 100 characters).
\end{itemize}
This domain-independent guideline can be an initial pathfinder to tailor the best examples in dialogue, though it may not be perfect.

\subsection{Intent Pool Construction Principles}
\label{app:principles_intent}
In constructing the intent labels, descriptions, and illustrative examples, we found that prompts yield the most generalizable and effective performance when they exhibit the following two key characteristics:
\begin{itemize}
    \item The intent list allows for mutually exclusive categorization of utterances.
    \item The explanations and examples clearly differentiate the degree of topic shifts induced by each utterance.
\end{itemize}
The intent specification adopted in Appendix~\ref{app:prompt_template} satisfies these conditions, and the intents tailored for other datasets (Appendix~\ref{app:prompt_other_labels}) were likewise designed to adhere to these principles.

\section{Analysis on other datasets}
\label{app:others}

\subsection{Ablation Study}
\label{app:others_ablation}

\begin{table}[htb]
  \tabcolsep=3.5pt
  \centering
  \small
    \resizebox{0.95\linewidth}{!}{
      \begin{tabular}{c|ccc|ccc}
        \toprule
          \multirow{2}{*}{\textbf{Method}} & 
          \multicolumn{3}{c|}{\textbf{SuperDialseg}} & 
          \multicolumn{3}{c}{\textbf{Dialseg711}} \\
          \cmidrule(lr){2-7} &
          \textbf{P$_k$$\downarrow$} & \textbf{WD$\downarrow$} & \textbf{F1$\uparrow$} & 
          \textbf{P$_k$$\downarrow$} & \textbf{WD$\downarrow$} & \textbf{F1$\uparrow$} \\
        \midrule
          w/o all & 
          0.378 & 0.382 & 0.467 & 
          0.007210 & 0.009416 & 0.987245 \\
          w/o intent & 
          0.363 & 0.364 & 0.448 & 
          0.093486 & 0.127211 & 0.701330 \\
          w/o examples & 
          0.338 & 0.341 & 0.646 & 
          0.005826 & 0.012322 & 0.984733 \\
          w/o context & 
          0.327 & 0.331 & 0.635 & 
          \textbf{0.005800} & \textbf{0.008464} & \textbf{0.989770} \\
          Def-DTS & 
          \textbf{0.317} & \textbf{0.322} & \textbf{0.674} & 
          0.009024 & 0.013738 & 0.982143 \\
        \bottomrule
      \end{tabular}
    }
    \caption{Ablation study for identifying effectiveness of each subtask within our method.}
    \label{tab:ablation_study_other_datasets}
\end{table}

As shown in Table~\ref{tab:ablation_study_other_datasets}, we found that the absence of some module leads to performance degradation. For efficient evaluation, we used 100 randomly sampled dialogue for the ablation study for SuperDialseg and Dialseg711 each. This data is the same as we used in Section~\ref{sec:analysis4}.
Especially on the SuperDialseg dataset, we observed a consistent improvement by adding any subtask.
However, with respect to Dialseg711 dataset, the existence of context extraction module is crucial to performance improvement. Even the w/o case surpasses our full-attached method. we conjecture that the issue is due to over-concentration for local context, same as the result of ablation study for TIAGE, moreover, on the case of dialseg711 having clear topic shift signal, just predicting label is enough to solve the problem.
After all, our intent classification module elevates the performance across all datasets.

\subsection{Intent Classification Accuracy}
\label{app:others_intent}

\begin{table}[htb]
  \centering
  \normalsize
  \renewcommand{\arraystretch}{1.05}
    \resizebox{1.00\columnwidth}{!}{
      \begin{tabular}{|c|cccc|c|}
        \toprule
        %   \textbf{Intent} & TP & FP & TN & FN & Acc \\
        % \midrule
          \multicolumn{6}{|c|}{\textbf{SuperDialseg}} \\
        \midrule
          \textbf{Intent} & TP & FP & TN & FN & Acc \\
        \midrule
          DIFFERENT\_QUESTION & 2456 & 688 & 83 & 192 & 0.74 \\
          RELEVANT\_QUESTION & 1 & 0 & 811 & 989 & 0.45\\
          ANSWERING & 0 & 2 & 7819 & 168 & 0.98 \\
          ADDITIONAL\_COMMENT & 0 & 0 & 1264 & 211 & 0.86 \\
        \midrule
          \multicolumn{6}{|c|}{\textbf{Dialseg711}} \\
        \midrule
          \textbf{Intent} & TP & FP & TN & FN & Acc \\
        \midrule
          JUST\_COMMENT & 0 & 6 & 5067 & 14 & 0.996 \\
          JUST\_ANSWER & 0 & 6 & 7675 & 8 & 0.998 \\
          DEVELOP\_TOPIC & 0 & 3 & 2359 & 13 & 0.993 \\
          CHANGE\_TOPIC & 2708 & 66 & 0 & 0 & 0.976 \\
        \bottomrule
      \end{tabular}
    }
    \caption{Intent-level confusion matrix for other datasets.}
    \label{tab:n_others_confusion_matrix}
\end{table}

We conducted a detailed accuracy analysis for the other datasets and the results are presented in Table~\ref{tab:n_others_confusion_matrix}.

For SuperDialseg, as shown in Table~\ref{tab:utterance_intents_superseg}, four new intent pools were applied. ADDITIONAL\_COMMENT, ANSWERING, and RELEVANT\_QUESTION are classified as non-topic shift cases, whereas the case of DIFFERENT\_QUESTION is classified as a topic shift case. However, the instruction following was not well executed for the case of DIFFERENT\_QUESTION. For the case of RELEVANT\_QUESTION, the following instruction was well executed with one exception, but its accuracy was relatively low. The difference in explanations between the case of RELEVANT\_QUESTION and DIFFERENT\_QUESTION could be linked to the actual dataset characteristics and topic changes. 
In contrast, the cases of ANSWERING and ADDITIONAL\_COMMENT showed significantly high classification accuracy. This comparison suggest that improving Question-type intents  will lead to an overall improvement in performance.

For Dialseg711, overall accuracy was higher compared to other datasets where the following deductive instruction not executed for less than 1\% of utterances. For the results of the three intents, excluding the case of CHANGE\_TOPIC, 18\% of DEVELOP\_TOPIC, 30\% of JUST\_COMMENT, and 42\% of JUST\_ANSWER cases among all false cases were misclassified due to errors in instruction following. It is believed that this issue can be resolved through additional instructions or prompt modifications for instruction following.

\section{Additional Experiments}
\label{app:additional_exps}

\subsection{Experiments for Local LLMs}
\label{app:local_llms}

\begin{table*}[thb]
  \tabcolsep=3.5pt
  \centering
  \small
  \renewcommand{\arraystretch}{1.05}
  \resizebox{\linewidth}{!}{
      \begin{tabular}{|l|l|cccc|cccc|cccc|}
        \toprule
          \multirow{2}{*}{\textbf{Method}} & 
          \multirow{2}{*}{\textbf{Model}} & 
          \multicolumn{4}{c|}{\textbf{TIAGE}} & 
          \multicolumn{4}{c|}{\textbf{SuperDialseg}} & 
          \multicolumn{4}{c|}{\textbf{Dialseg711}} \\
        \cmidrule(lr){3-6}\cmidrule(lr){7-10}\cmidrule(lr){11-14}
          & & \textbf{P$_k$$\downarrow$} & \textbf{WD$\downarrow$} & \textbf{F1$\uparrow$} & \textbf{Error} & \textbf{P$_k$$\downarrow$} & \textbf{WD$\downarrow$} & \textbf{F1$\uparrow$} & \textbf{Error} & \textbf{P$_k$$\downarrow$} & \textbf{WD$\downarrow$} & \textbf{F1$\uparrow$} & \textbf{Error} \\
    \bottomrule
    \midrule
      \multirow{3}{*}{Plain Text} & Llama 8B & 
      0.529 & 0.604 & 0.303 & 1 & 
      0.497 & 0.504 & 0.036 & 0 & 
      0.350 & 0.373 & 0.032 & 0 \\
       & Qwen 7B & 
      0.509 & 0.563 & 0.249 & 2 & 
      0.517 & 0.522 & 0.132 & 1 & 
      0.486 & 0.513 & 0.069 & 0 \\
       & Qwen 32B & 
      0.476 & 0.515 & 0.221 & 0 & 
      0.466 & 0.471 & 0.083 & 0 & 
      0.391 & 0.415 & 0.015 & 0 \\
    \midrule
      \multirow{3}{*}{S3-DST\textsubscript{uttr}} & Llama 8B & 
      0.460 & 0.460 & 0.018 & 0 & 
      0.472 & 0.494 & 0.076 & 0 & 
      0.188 & 0.196 & 0.705 & 0 \\
       & Qwen 7B & 
      0.563 & 0.860 & 0.299 & 57 & 
      0.578 & 0.952 & 0.351 & 47 & 
      0.582 & 0.759 & 0.093 & 71 \\
       & Qwen 32B & 
      0.430 & 0.455 & 0.211 & 22 & 
      0.431 & 0.443 & 0.106 & 57 & 
      0.237 & 0.270 & 0.497 & 83 \\
    \midrule
      \multirow{3}{*}{Def-DTS} & Llama 8B & 
      0.474 & 0.525 & 0.218 & 1 & 
      0.473 & 0.527 & 0.398 & 0 & 
      0.246 & 0.268 & 0.464 & 10 \\
       & Qwen 7B & 
      0.462 & 0.465 & 0.022 & 20 & 
      0.473 & 0.474 & 0.071 & 14 & 
      0.162 & 0.170 & 0.715 & 44 \\
       & Qwen 32B & 
      0.338 & 0.374 & 0.501 & 38 & 
      0.392 & 0.400 & 0.429 & 1 & 
      0.221 & 0.343 & 0.435 & 18 \\
    \bottomrule
  \end{tabular}
  }
  \caption{Performances on the local LLMs. We employed Llama-3.1-8B-Instruct, Qwen2.5-7B-Instruct and Qwen2.5-32B-Instruct for Llama 8B, Qwen 7b and Qwen 32B, respectively. P$_k$, WD, F1 were calculated only for correctly formatted outputs. }
  \label{tab:app_local_llms}
\end{table*}

Def-DTS leverages LLM reasoning capabilities, so model size significantly affects performance. Table~\ref{tab:main_local_llms} shows that S3-DST on Qwen 70B had formatting errors, which discouraged us from testing smaller LLMs initially. However, considering the growing capabilities and applications of sLLMs, we conducted additional experiments on: Llama 8B, Qwen 7B, Qwen 32B. 

The experimental result is shown in table~\ref{tab:app_local_llms}. Although Def-DTS struggled with smaller models and dialseg711, it showed greater improvements with larger models by leveraging LLM reasoning capability. However, we acknowledge that applying Def-DTS to smaller LLMs would require adjustments such as additional parameter modification.

\subsection{Performance for Closed-source LLMs}
\label{app:closed_llms}

\begin{table}[htb]
  \centering
  \tiny
    \resizebox{0.95\linewidth}{!}{
      \begin{tabular}{cccc}
        \toprule
          \multirow{2}{*}{\textbf{Models}}  & \multicolumn{3}{c}{\textbf{TIAGE}} \\
          \cmidrule(lr){2-4}
          & \textbf{P$_k$$\downarrow$} & \textbf{WD$\downarrow$} & \textbf{F1$\uparrow$} \\
        \midrule
          R1 & 0.286 & 0.331 & 0.644 \\
          V3 & 0.259 & 0.204 & 0.674 \\
          GPT-4o & 0.232 & 0.256 & 0.699 \\
        \bottomrule
      \end{tabular}
    }
    \caption{Performance for additional closed-source LLMs.}
    \label{tab:app_deepseek}
\end{table}

We evaluate the performance of Def-DTS with additional closed-source LLMs, including Deepseek-R1\citep{deepseek_r1_2025} and Deepseek-V3\citep{deepseep_v3_2024}, as shown in Table~\ref{tab:app_deepseek}.
All three models achieve better performance than the LLM-based approaches reported in Section~\ref{sec:results}, and—with the exception of R1—even outperform the supervised baseline. This demonstrates the general applicability of Def-DTS across a variety of closed-source LLMs.
Interestingly, although R1 is specialized for reasoning, it underperforms compared to GPT-4o and V3, which are not explicitly designed for such capabilities. We attribute this to our prompting strategy: by providing explicit topic-shift criteria, illustrative examples, and a clearly defined reasoning path, the task is structured in a way that reduces the need for complex, proactive reasoning.

\section{Details for Experiment}
\label{app:details}

\subsection{Details for Implementation}
\label{app:details_imp}

The model used for our experiments is \href{https://openai.com/index/hello-gpt-4o/}{gpt-4o} for closed LLM, \href{https://huggingface.co/meta-llama/Llama-3.1-70B-Instruct}{Llama-3.1-70B-Instruct} and \href{https://huggingface.co/Qwen/Qwen2.5-72B-Instruct}{Qwen2.5-72B-Instruct} for open-source LLM. At first, we considered two closed models: gpt-4o and \href{https://www.anthropic.com/news/claude-3-5-sonnet}{Claude-3.5-sonnet}. But Claude was excluded due to poor accuracy compared to gpt-4o at a preliminary evaluations and there are no prior studies applied their method to the claude family. For the inference of open-source LLMs, we utilized a computational infrastructure consisting of 4*NVIDIA A100 80GB GPU.  We conducted our experiments without employing any model-specific tuning or quantization techniques, thus maintaining the original model architecture and parameters. We kept the hyperparameters initially stated, except for the temperature that we set to 0 for reproducibility of our experiments.

\subsection{Details for Reproduce}
\label{app:details_rep}

\begin{table}[thb]
  \tabcolsep=3.5pt
  \centering
  \normalsize
  \renewcommand{\arraystretch}{1.05}
    \resizebox{0.95\linewidth}{!}{
      \begin{tabular}{|l|ccc|ccc|ccc|}
        \toprule
          \multirow{2}{*}{\textbf{Model}}  & \multicolumn{3}{c|}{\textbf{TIAGE}} & \multicolumn{3}{c|}{\textbf{SuperDialseg}} & \multicolumn{3}{c|}{\textbf{Dialseg711}} \\
        \cmidrule(lr){2-4}\cmidrule(lr){5-7}\cmidrule(lr){8-10}
          & \textbf{P$_k$$\downarrow$} & \textbf{WD$\downarrow$} & \textbf{F1$\uparrow$} & \textbf{P$_k$$\downarrow$} & \textbf{WD$\downarrow$} & \textbf{F1$\uparrow$} & \textbf{P$_k$$\downarrow$} & \textbf{WD$\downarrow$} & \textbf{F1$\uparrow$} \\
        % \cmidrule(lr){2-16}
        % \midrule
        \bottomrule
        \toprule
          \multicolumn{10}{|c|}{\textbf{Unsupervised Learning Methods}} \\
        \midrule
          DialSTART & 
          - & - & - & 
          - & - & - & 
          0.179 & 0.198 & - \\
          SumSeg & 
          0.438 & 0.455 & - & 
          0.469 & 0.480 & - & 
          - & - & - \\
        \bottomrule
        \toprule
          \multicolumn{10}{|c|}{\textbf{LLM-based Methods}}\\
        \midrule
          Plain Text (GPT-3.5) & 
          0.496 & 0.560 & 0.362 & 
          0.318 & 0.347 & 0.658 & 
          0.290 & 0.355 & 0.690 \\
          S3-DST$_{turn}$ & 
          - & - & - & 
          - & - & - & 
          0.009 & 0.008 & - \\
        \bottomrule
      \end{tabular}
      }
  \caption{Performances reported in their original papers. Denoted as DialSTART indicates main result of \citealp{dialstart_2023}, SumSeg indicates main result of \citealp{sumseg_2024}, Plain Text indicates ChatGPT variant of \citealp{superdialseg_2023}'s main result and S3-DST$_{turn}$ indicates main result of \citealp{s3_dst_2024}.}
  \label{tab:other_result}
\end{table}

For the SuperDialseg paper\citep{superdialseg_2023}, we obtained experimental results for the Random, TextTiling, TextTiling+Glove, CSM, BERT, RoBERTa, and RetroTS-T5 methods.

We reproduced the experimental result for PlainText\citep{superdialseg_2023}, S3-DST\citep{s3_dst_2024}, DialSTART\citep{dialstart_2023}, and SumSeg\citep{sumseg_2024}, which either lacked F1 scores or did not perform experiments on certain datasets. 

During reproduction, we maintained all settings, including seeds, without any parameter modifications. However, We observed results that differed from the original experiments. Their original experimental results are presented in Table~\ref{tab:other_result}.

For the reproduction of LLM-based methods, we made the necessary modifications in the following cases. 

Plain Text \citep{superdialseg_2023}: As Plain Text was the only methodology that disclosed the system prompt, we equitably refrained from using system prompts. We compared results with and without system prompts, finding nearly identical performance aside from parsing inconveniences.

S3-DST \citep{s3_dst_2024}: S3-DST constructs prompts on a turn basis, while we performed on an utterance basis. Their approach is not suitable for our approach when dialogue has consecutive utterance or odd numbers of utterances. Therefore, we modified turn-based inference to utterance-based.

All final prompts used are attached to our repository. 

% \section{Licenses for artifacts}
% \paragraph{Datasets}

% TIAGE, SuperDialseg: MIT License,

% Dialseg711: We were unable to find a license for this dataset.
% However, you can find details of the dataset in the original paper\citep{xu2021topic}.

% \paragraph{Models}

% Qwen2.5-70B-Instruct, Qwen2.5-32B-Instruct, Qwen2.5-7B-Instruct: Qwen License,

% Llama3.1-72B-Instruct, Llama3.1-8B-Instruct: Llama License.

\end{document}